%% file: main.tex
\documentclass[10pt,twocolumn,letterpaper]{article}

\usepackage{cvpr}              

\usepackage{graphicx}
\usepackage{amsmath}
\usepackage{amssymb}
\usepackage{booktabs}
\usepackage{dsfont}
\usepackage{multirow}
\usepackage{algorithm}
\usepackage{algpseudocode}
\usepackage{xcolor}
\usepackage{colortbl}
\definecolor{graylite}{gray}{.89}

\usepackage[pagebackref,breaklinks,colorlinks]{hyperref}

\usepackage[capitalize]{cleveref}
\crefname{section}{Sec.}{Secs.}
\Crefname{section}{Section}{Sections}
\Crefname{table}{Table}{Tables}
\crefname{table}{Tab.}{Tabs.}

\def\cvprPaperID{10143} 
\def\confName{CVPR}
\def\confYear{2022}

\begin{document}

\title{Crafting Better Contrastive Views for Siamese Representation Learning}

\author{
Xiangyu Peng\textsuperscript{1}\thanks{Equal contribution.} 
\quad Kai Wang\textsuperscript{1}\footnotemark[1]
\quad Zheng Zhu\textsuperscript{2}
\quad Mang Wang\textsuperscript{3}
\quad Yang You\textsuperscript{1}\thanks{Corresponding author.}
\\
\textsuperscript{1}{National University of Singapore} 
\quad \textsuperscript{2}{Tsinghua University}
\quad \textsuperscript{3}{Alibaba Group}
\\
\small{\texttt{\{xiangyupeng, kai.wang, youy\}@comp.nus.edu.sg}} \\
\small{\texttt{zhengzhu@ieee.org}
\quad \texttt{wangmang.wm@alibaba-inc.com}}
\\
\small{Code: \url{https://github.com/xyupeng/ContrastiveCrop}}
}

\maketitle

\input{sec/abs}
\input{sec/intro}

\input{sec/related_v2}
\input{sec/method_v2}
\input{sec/exp_v2}

\input{sec/conclusion}

{\small
\bibliographystyle{ieee_fullname}
\bibliography{egbib}
}

\end{document}

%% file: sec/abs.tex
\begin{abstract}
Recent self-supervised contrastive learning methods greatly benefit from the Siamese structure that aims at minimizing distances between positive pairs. For high performance Siamese representation learning, one of the keys is to design good contrastive pairs. Most previous works simply apply random sampling to make different crops of the same image, which overlooks the semantic information that may degrade the quality of views. In this work, we propose \textit{ContrastiveCrop}, which could effectively generate better crops for Siamese representation learning. Firstly, a semantic-aware object localization strategy is proposed within the training process in a fully unsupervised manner. This guides us to generate contrastive views which could avoid most false positives (\ie, object \vs background). Moreover, we empirically find that views with similar appearances are trivial for the Siamese model training. Thus, a center-suppressed sampling is further designed to enlarge the variance of crops. Remarkably, our method takes a careful consideration of positive pairs for contrastive learning with negligible extra training overhead. As a plug-and-play and framework-agnostic module, \textit{ContrastiveCrop} consistently improves SimCLR, MoCo, BYOL, SimSiam by $0.4\% \sim 2.0\%$ classification accuracy on CIFAR-10, CIFAR-100, Tiny ImageNet and STL-10. Superior results are also achieved on downstream detection and segmentation tasks when pre-trained on ImageNet-1K.
\end{abstract}

%% file: sec/intro.tex
\section{Introduction}
\label{sec:intro}


\begin{figure}[htp]
    \centering
    \begin{subfigure}{0.23\textwidth}
        \includegraphics[width=\linewidth, trim=12cm 6cm 12cm 6cm, clip]{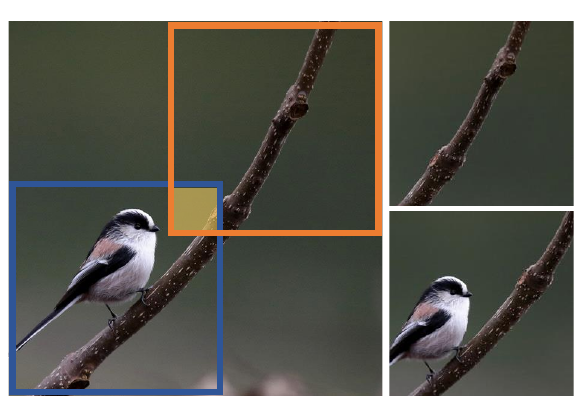}
        \caption{}
        \label{fig:moti_a}
    \end{subfigure}
    \begin{subfigure}{0.23\textwidth}
        \includegraphics[width=\linewidth, trim=12cm 6cm 12cm 6cm, clip]{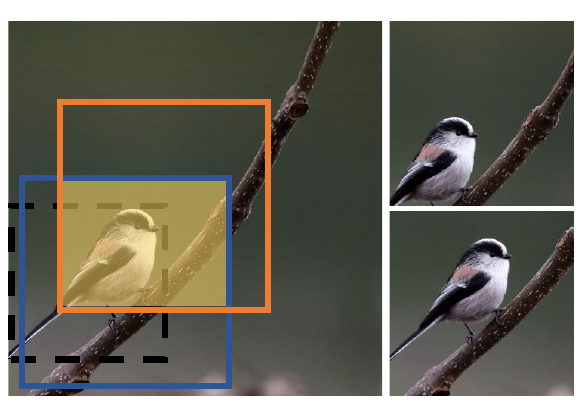}
        \caption{}
        \label{fig:moti_b}
    \end{subfigure}
    \begin{subfigure}{0.23\textwidth}
        \includegraphics[width=\linewidth, trim=12cm 6cm 12cm 6cm, clip]{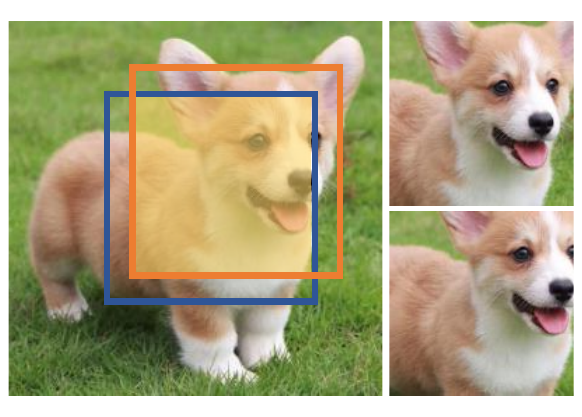}
        \caption{}
        \label{fig:moti_c}
    \end{subfigure}
    \begin{subfigure}{0.23\textwidth}
        \includegraphics[width=\linewidth, trim=12cm 6cm 12cm 6cm, clip]{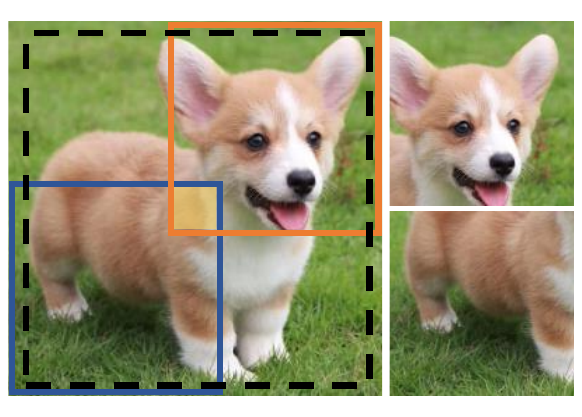}
        \caption{}
        \label{fig:moti_d}
    \end{subfigure}
\caption{The motivation of our proposed \textit{ContrastiveCrop}. (a) and (c) are generated by typical \textit{RandomCrop}, while (b) and (d) are crops from our method. We address the false positive problem (object \vs background) shown in (a) by localizing the object and restricting the crop center within the bounding box (the black dashed box)  in (b). Moreover, we enlarge the variance of crops in (d) by keeping them away from the center, which avoids the close appearance as shown in (c).}
\label{motivation}
\end{figure}

Self-supervised learning (SSL) has attracted much attention in the computer vision community due to its potential of exploiting large amount of unlabeled data. As a mainstream approach in SSL, contrastive learning has achieved higher performance on several downstream tasks (\eg, object detection, segmentation and pose estimation \cite{faster, he2017mask, guler2018densepose, everingham2010pascal_VOC, lin2014microsoft_coco, zhang2021learning}) than its supervised counterpart. Such promising results can be largely attributed to the Siamese structure, which is commonly applied in state-of-the-art unsupervised methods, including SimCLR \cite{chen2020simple}, MoCo V1 \& V2 \cite{he2020momentum, chen2020improved}, BYOL \cite{grill2020bootstrap} and SimSiam \cite{chen2020exploring}. Typically, the Siamese structure takes two augmented views from an image as input, and minimizes their distance in the embedding space. With proper views selected, Siamese networks demonstrate a strong capability to learn generic visual features \cite{tian2020makes}.

One of the key issues of contrastive learning is to design positives selection. Some works generate different positive views by strong data augmentation, such as color distortion and jigsaw transformation \cite{tian2020makes, chen2021jigsaw}. Another work \cite{shen2020mix} applies mixture \cite{zhang2017mixup, yun2019cutmix} in an unsupervised manner to produce positive pairs that incorporate multiple samples. Additionally, different from data augmentation, \cite{zhu2021improving} creates hard positives with transformation at the feature level. Despite different techniques, these works commonly apply \textit{RandomCrop} to sample multiple views of an image, and further make the views more diverse. 

As a basic sampling method, \textit{RandomCrop} enables all individual crops to be selected equiprobably. However, it fails to look at the semantic information of paired views, which helps to learn better representations more efficiently and accurately. As shown in Fig.~\ref{fig:moti_a}, random crops are prone to miss the object when no prior of object (\eg, scale and location) is given. Optimizing the distance between object and background in the embedding space would mislead the learning of representations. Besides, Fig.~\ref{fig:moti_c} indicates that random crops cannot always carry sufficient variances of an object. Such views with large similarity are trivial for learning discriminative models.

In this paper, we propose \textit{ContrastiveCrop}, aiming to craft better contrastive pairs for Siamese representation learning. False positives indicate that a better sampling strategy for contrastive learning should consider the content information of the image. Hereby, we propose a semantic-aware localization scheme. The module serves as a guidance to select crops, avoiding most false positives, as shown in Fig.~\ref{fig:moti_b}. Moreover, we propose a center-suppressed sampling strategy to tackle trivial positive pairs with large similarity. Fig.~\ref{fig:moti_d} shows that our crops are more likely to cover different parts of the object. The semantic-aware localization and center-suppressed sampling scheme can be gracefully combined to generate better crops for contrastive learning.

The proposed \textit{ContrastiveCrop} considers both semantic information and maintaining large variance when making pairs. As a plug-and-play method, it can be easily applied into the Siamese structure. More importantly, our approach is agnostic to contrastive frameworks, regardless using negative samples or not. With negligible training overhead, our strategy consistently improves \mbox{SimCLR}, MoCo, BYOL, SimSiam by $0.4\% \sim 2.0\%$ classification accuracy on \mbox{CIFAR-10}, CIFAR-100, Tiny ImageNet and STL-10. Superior results are also achieved on downstream detection and segmentation tasks when pre-trained on ImageNet-1K.

The main contributions of this paper can be summarized as:
\begin{itemize}
    \item To the best of our knowledge, this is the first work to investigate the problem of commonly used \textit{RandomCrop} in contrastive learning. We propose \textit{ContrastiveCrop} that is customized to generate better views for this task.
    \item In \textit{ContrastiveCrop}, the semantic-aware localization is adopted to avoid most false positives and the center-suppressed sampling strategy is applied to reduce trivial positive pairs.
    \item \textit{ContrastiveCrop} consistently outperforms \textit{RandomCrop} with popular contrastive methods on a variety of datasets, showing its effectiveness and generality for Siamese representation learning.
\end{itemize}

%% file: sec/related_v2.tex
\section{Related works}
\label{sec:related}

In this section, we introduce contrastive learning and positives selection related to this work.

\subsection{Contrastive Learning}
The core idea of contrastive learning is to pull positive pairs closer while pushing negatives apart in the embedding space. This methodology has shown great promise in learning visual representations without annotation \cite{bachman2019learning, henaff2019data_CPCv2, wu2018unsupervised, misra2020self_PIRL, oord2018representation, ye2019unsupervised, tian2019contrastive}. More recently, contrastive methods based on the Siamese structure achieve remarkable performance on downstream tasks \cite{chen2020simple, he2020momentum, chen2020improved, grill2020bootstrap, chen2020exploring, xie2021detco, wang2020dense, xie2020propagate, dwibedi2021little}, some of which even surpass supervised models.

The milestone work is SimCLR \cite{chen2020simple}, which presents a simple framework for contrastive visual representation learning. It significantly improves the quality of learned representations with a non-linear transformation head. Another famous work is MoCo \cite{he2020momentum}, which uses a memory bank to store large number of negative samples and smoothly updates it with momentum for better consistency. Methods that learn useful representations without negative samples are also proposed. BYOL \cite{grill2020bootstrap} trains an online network to predict the output of the target network, with the latter slowly updated with momentum. The authors hypothesize that the additional projector to the online network and the momentum encoder are important to avoid collapsed solutions without negative samples. SimSiam \cite{chen2020exploring} further explores simple Siamese networks that can learn meaningful representations without negative sample pairs, large batches and momentum encoders. The role of stop-gradient is emphasized in preventing collapsing. In addition to framework design, theoretical analyses and empirical studies have also been proposed to better understand the behavior and properties of contrastive learning \cite{wang2020understanding, wang2020understanding, zoph2020rethinking, xiao2020should, wu2020mutual, arora2019theoretical, hua2021feature, purushwalkam2020demystifying, tian2021understanding, chen2020intriguing, chen2021empirical, caron2021emerging}. 

\begin{figure*}[ht] 
\centering
\includegraphics[width=\textwidth, trim=1cm 6cm 1cm 6cm, clip]{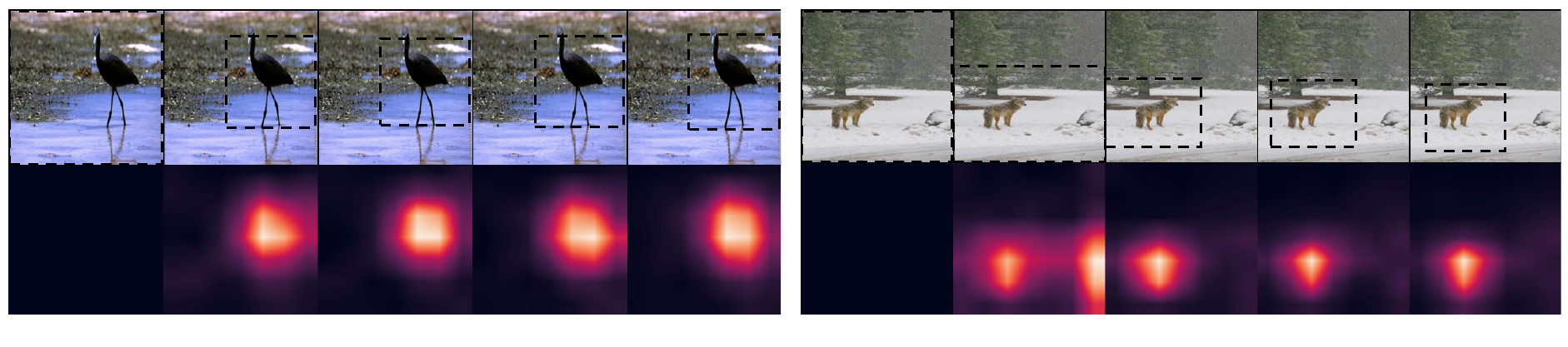}
\caption{The training dynamic of localization is shown from left to right in each subfigure. We initialize the localization box as the whole image, and update it at a regular interval using the latest heatmap. Note that our goal is not to derive precise localization, but to guide generation of crops by finding the object of interest.}
\label{fig:boxes}
\end{figure*}

\subsection{Positives Selection}
One of the key issues in contrastive learning is the design of positives selection. An intuitive approach to generating positive pairs is to create different views of a sample using data augmentation. Most SSL works apply data augmentation pipelines that are directly adapted from those in supervised learning \cite{cubuk2020randaugment, cubuk2018autoaugment, lim2019fast, hataya2020faster, zhang2017mixup, yun2019cutmix}. Chen~\etal \cite{chen2020simple} comprehensively study the effect of a range of data transformations, and find out the composition made of random cropping and random color distortion can lead to better performance. Tian~\etal \cite{tian2020makes} propose an \textit{InfoMin principle} to catch a sweet point of mutual information between views, and accordingly generate positive pairs with its \textit{InfoMin Augmentation}. A close work to this paper is~\cite{Selvaraju_2021_CVPR}, which also uses unsupervised saliency maps as a constraint of crops, but crops are still randomly sampled. All these works commonly apply \textit{RandomCrop} as the basic sampling method to generate input views, which we find may not be the optimal solution for contrastive learning. \cite{mishra2021object} take object-scene relation into account when making crops, but require additional object proposal algorithms. In this work, we propose \textit{ContrastiveCrop} that is tailored to create better positives views for contrastive learning, without the need of external functions. 

%% file: sec/method_v2.tex
\section{Method}

In this section, we introduce \textit{ContrastiveCrop} for Siamese representation learning. Firstly, we briefly review
\textit{RandomCrop} as the preliminary knowledge. Then, we describe semantic-aware localization and center-suppressed sampling as two submodules of our \textit{ContrastiveCrop}. Finally, favorable properties of our method are further discussed for better understanding.

\subsection{Preliminary}
\textit{RandomCrop}, an efficient data augmentation method, has been widely used in both supervised learning and self-supervised learning (SSL). Here, we briefly review this technique, using API in Pytorch$\footnote{https://pytorch.org/vision/stable/transforms.html}$ as an example. Given an image $I$, we first determine the scale $s$ and aspect ratio $r$ of the crop from a pre-defined range (\eg,  $s \in [0.2, 1.0]$ and $r \in [3/4, 4/3]$) Then, the height and width of the crop can be obtained with $s$ and $r$. After that, the location of the crop is randomly selected on the image plane, as long as the whole crop lies within the image. The procedure of \textit{RandomCrop} can be formulated as
\begin{equation}
    (x, y, h, w) = \mathbb{R}_{crop}(s, r, I),
    \label{eq:1}
\end{equation}
where $\mathbb{R}_{crop}(\cdot, \cdot, \cdot)$ is the random sampling function that returns a quaternion $(x,y,h,w)$ representing the crop. We denote $I$ as the input image, $(x, y)$ as the coordinate of the crop center, and ($h$, $w$) as the height and width of crop. Usually, the scale $s$ and aspect ratio $r$ of crops are set flexibly, so that crops of variant sizes could be made.

\begin{figure*}[htp] 
\centering
    \includegraphics[width=\textwidth, trim=3.8cm 2cm 4.5cm 2.3cm, clip]{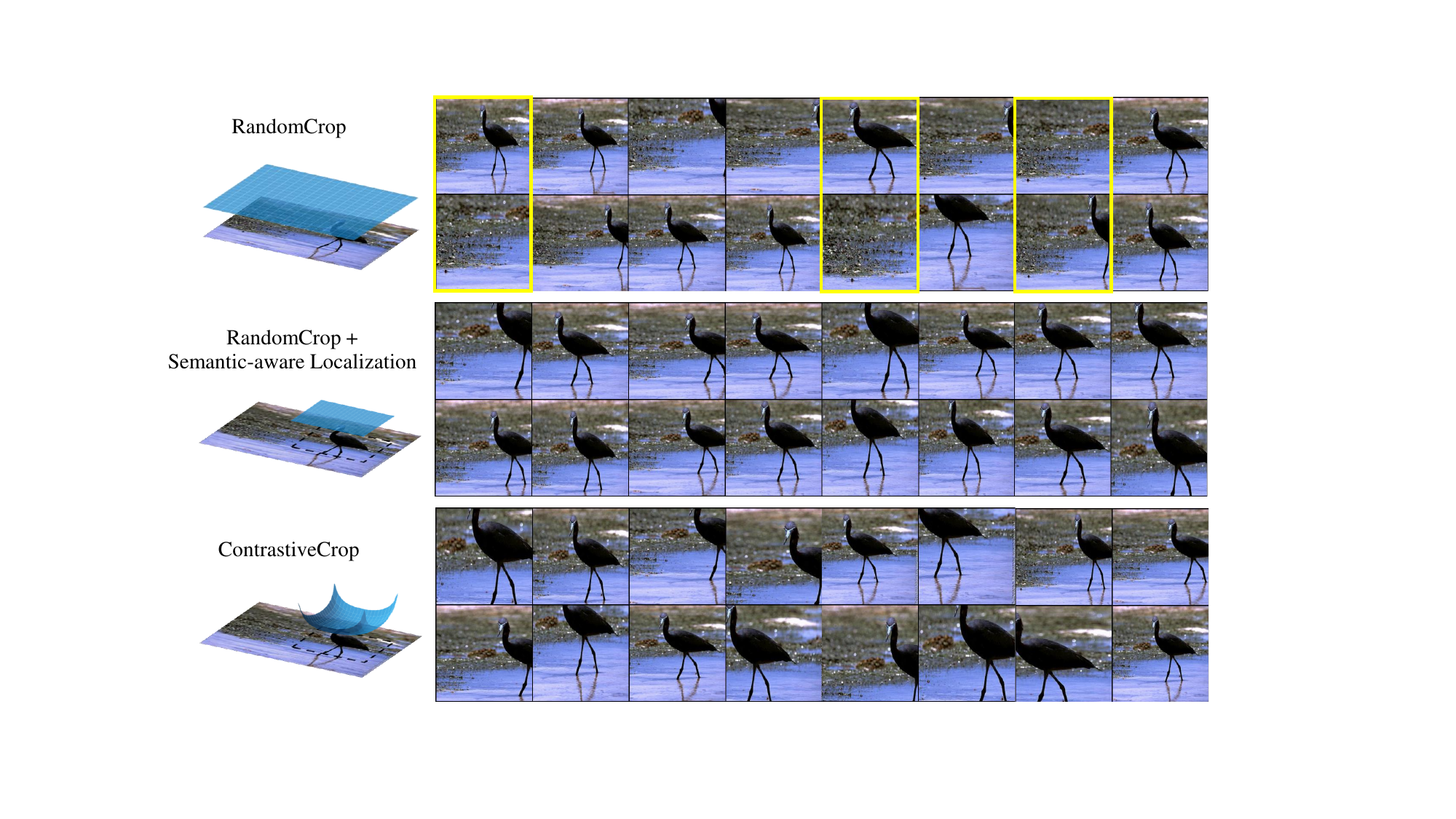}
\caption{Visualization of \textit{RandomCrop}, \textit{RandomCrop + Semantic-aware Localization} and our \mbox{\textit{ContrastiveCrop}}. We show the sampling distributions and operable regions for three settings on the left, and correspondent sampled pairs on the right. Pairs made by \textit{RandomCrop} include several false positives that totally miss the object (marked in yellow box). Using \textit{RandomCrop} with \textit{Semantic-aware Localization} reduces false positives, but introduces easy positive pairs that share large similarity. Last, our \textit{ContrastiveCrop} could reduce false positive pairs while increasing variance at the same time. }
\label{fig:views}
\end{figure*}

In principle, \textit{RandomCrop} enables all individual crops to be selected, thus could provide diverse views of a sample. However, it performs sampling equiprobably (\ie, each single view is sampled with the same probability), which ignores the semantic information of images. As shown in Fig.~\ref{fig:moti_a}, \textit{RandomCrop} is prone to generate false positives when the scale of object is small. Given objects with variant scales in contrastive learning, \textit{RandomCrop} would inevitably generate false positives due to lack of the consideration of semantic information. As a result, optimizing the false positives in Fig.~\ref{fig:views} may mislead the learning of good representations. Therefore, designing a semantic-aware sampling strategy for crops is crucial and vital for Siamese representation learning.

\subsection{Semantic-aware Localization}  \label{sec:box}
To tackle the issue of poor content understanding in \textit{RandomCrop}, we design a semantic-aware localization module that can effectively reduce false positives in an unsupervised manner. To better study the process of feature learning in Siamese networks, we visualize the heatmaps generated at different training stages (\eg, 0\textit{th}, 20\textit{th}, 40\textit{th}, 60\textit{th}, 80\textit{th} epoch) in Fig.~\ref{fig:boxes}. Note that we derive the heatmap by summing the features of last convolutional layer across the channel dimension and normalizing it to [0, 1]. There are several inspirations from visualization: 1) The Siamese representation learning framework is capable of capturing the location of the object, which can be leveraged to guide the generation of better crops; 2) Heatmaps can roughly indicate the object, but may need some warm-up at early stages.

Based on above analyses, we propose to locate the object during the training process using the information in heatmaps. Specifically, \textit{RandomCrop} is applied at early stage of training to collect semantic information of the whole image. Then, we apply an indicator function to obtain the bounding box of object $B$ from heatmaps, which can be written as,
\begin{equation} \label{eq:bb}
    B = L(\mathds{1}[M > k]),
\end{equation}
where $M$ represents heatmap, $k \in [0, 1]$ is the threshold of activations, $\mathds{1}$ is the indicator function and $L$ calculates the rectangular closure of activated positions. After obtaining the bounding box $B$, the semantic crops could be generated as follows,
\begin{equation}
    (\dot{x}, \dot{y}, \dot{h}, \dot{w}) = \mathbb{R}_{crop}(s, r, B),
    \label{eq:3}
\end{equation}
where the definitions of $\dot{x}$, $\dot{y}$, $\dot{h}$, $\dot{w}$, $s$, $r$, and $\mathbb{R}_{crop}$ are similar to Eq.~\ref{eq:1}. Considering the probable coarse localization, we enlarge the operable region by only constraining center of crops within $B$. This also reduces the potential negative effect of resolution discrepancy at training and inference stages \cite{touvron2019fixing}. 

At the training stage, the bounding box is progressively updated at a regular interval to leverage the latest features learned by the model. Note that our goal is not to derive precise localization, but to guide generation of crops by finding the object of interest. The scale of the bounding box is controlled by the threshold parameter $k \in [0,1]$. Generally speaking, a large $k$ leads to a small box and would limit the diversity of possible crops to be made. A small $k$, however, may still include much unrelated background texture and is not sufficient for finding the object. We study the effect of different threshold $k$ in Sec.~\ref{sec:abl}. We empirically find that the proposed localization module is not sensitive to this parameter and could improve over baseline within a wide range of $k$.

Finally, we show the sampling effect of semantic-aware localization in Fig.~\ref{fig:views}. Compared with \textit{RandomCrop}, one can find that the false positive pairs reduce dramatically when the proposed module is applied. This provides evidence that self-supervised neural networks trained without annotations are capable of recognizing the object of interest as well as its location. In this way, additional region proposals or ground truth bounding boxes are no longer necessary for views generation \cite{cheng2014bing, zitnick2014edge}. 

\subsection{Center-suppressed Sampling}

The semantic-aware localization scheme provides useful guidance to reduce false positive cases, but increases the probability of close appearance pairs due to the smaller operable region. In this subsection, we introduce the center-suppressed sampling that aims to tackle this dilemma. 

The main idea is to reduce the probability of crops gathering around center by pushing them apart. Specifically, we adopt the beta distribution $\beta(\alpha, \alpha)$ with two identical parameters $\alpha$, which shows a symmetric function. In this way, we could easily control the shape of the distribution with different $\alpha$. As the goal is to enlarge the variance of crops, we set $\alpha < 1$ which gives us a U-shaped distribution (\ie, with lower probability near center and greater one at other positions). In this way, crops are more likely to be scattered to near border lines of the operable region, and cases of much overlap could be largely avoided.

Combining center-suppressed sampling with semantic-aware localization, we can finally formulate our \textit{ContrastiveCrop} as
\begin{equation}
    (\dot{x}, \dot{y}, \dot{h}, \dot{w}) = \mathbb{C}_{crop}(s, r, B),
\end{equation}
where $\mathbb{C}_{crop}$ denotes sampling function that applies a center-suppressed distribution, and $B$ is the same bounding box as in Eq.~\ref{eq:3}. Note that the shape of beta distribution is determined by the parameter $\alpha$ and affects the variance of crops. We study the impact of different $\alpha$ in in Sec.~\ref{sec:abl}, including $\alpha > 1$ that gives a inverted U shape.

\begin{algorithm}
\caption{\textit{ContrastiveCrop} for Siamese Representation Learning}
\label{alg:code}
\begin{algorithmic} 
\State \textbf{Input:} Image $I$, Crop Scale $s$, Crop Ratio $r$, Threshold of Activations $k$, Parameter of $\beta$ Distribution $\alpha$.
\State $h = \sqrt{s \cdot r}$ \Comment{Height of the crop}
\State $w = \sqrt{s / r}$ \Comment{Width of the crop}
\State $F = Forward(I)$ \Comment{Features of last layer}
\State $M = Normalize(F)$ \Comment{Heatmap after normalizing}
\State $B = L(\mathds{1}[M > k])$ \Comment{Bounding box by Eq.~\ref{eq:bb}.} 
\State $x= B_{x0} + (B_{x1} - B_{x0}) \cdot u, u \sim \beta(\alpha, \alpha)$ 
\State $y= B_{y0} + (B_{y1} - B_{y0}) \cdot v, v \sim \beta(\alpha, \alpha)$ \\
\Comment{Sample crop center $x$ and $y$ from $\beta$ distribution}
\State \textbf{Output:} Crop $C=(x, y, h, w)$
\end{algorithmic}
\end{algorithm}

\begin{figure}[htp]
    \centering
    \includegraphics[width=0.42\textwidth, trim=0cm 0cm 1cm 1cm, clip]{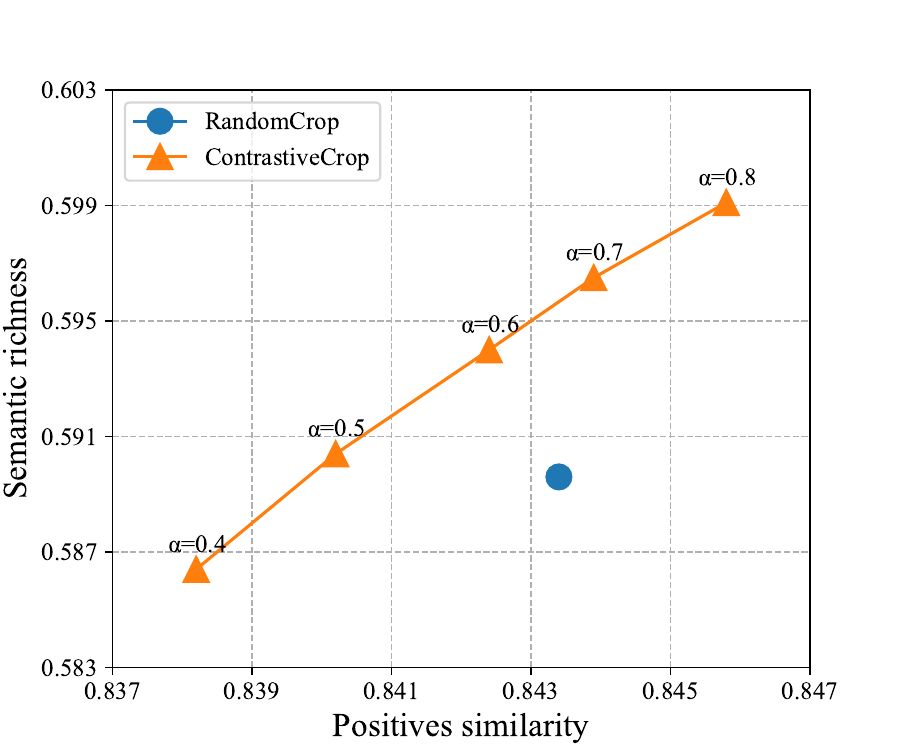}
\caption{The relation between semantic richness and positives similarity. Dots are obtained by varying $\alpha$ (fix $k=0.1$, delineated in Sec.~\ref{sec:impl}), and scores of each dot are calculated by averaging results of a large number of cropping trials. Compared with \textit{RandomCrop}, our \textit{ContrastiveCrop} conveys more semantic information at the same level of similarity (vertical), and yields less similar positive pairs under equal semantic information (horizontal).}
\label{fig:sem_var}
\end{figure}

\begin{figure}[htp]
    \centering
    \includegraphics[width=0.42\textwidth, trim=0cm 0cm 1cm 1cm, clip]{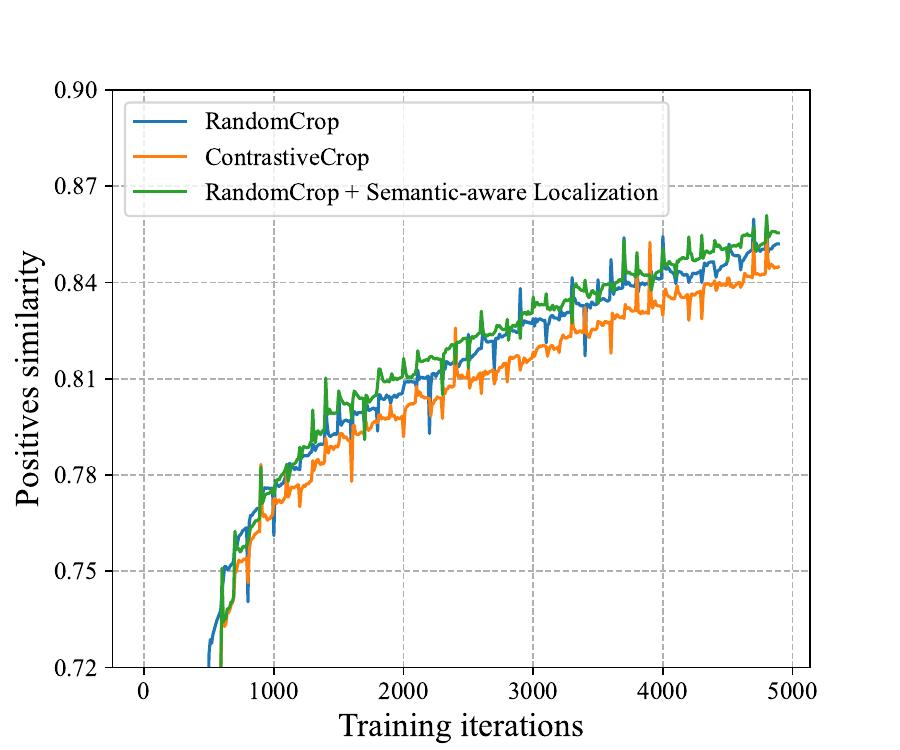}
\caption{Similarity of positive pairs in training. Smaller positives similarity indicates harder positive samples which may enhance representation learning \cite{zhu2021improving}. Taking \textit{RandomCrop} as baseline, adding only localization results in slightly larger similarity. Our \textit{ContrastiveCrop} combines both semantic-aware localization and center-suppressed sampling, which effectively reduces similarity of positives.}
\label{fig:pos_sim}
\end{figure}

The effect of our \textit{ContrastiveCrop} is visualized in Fig.~\ref{fig:views}. Compared with \textit{RandomCrop}, our method could significantly reduce false positive pairs due to the semantic-aware localization. Meanwhile, it introduces larger variance within a positive pair by applying the center-suppressed distribution. We show the pipeline for \textit{ContrastiveCrop} in Algorithm \ref{alg:code}. The whole module is agnostic to other transformations and can be easily integrated into general contrastive learning frameworks.


\input{tab/tab_non_IN}

\subsection{Discussion}
To better understand the behavior of \textit{ContrativeCrop}, we discuss several properties that may contribute to its effectiveness. We first investigate the relation between semantic information and positives similarity. We take the class score of a crop as an indicator of richness of categorized semantic information. The similarity of positive pairs is calculated in the latent space as the cosine similarity between positive representations. Both the class score and similarity are average results of a large number of cropping trials from a standard ResNet-50~\cite{res} trained with ImageNet~\cite{deng2009imagenet} labels. Their relation is shown in Fig.~\ref{fig:sem_var}. One can find that \textit{ContrastiveCrop} conveys more semantic information than \textit{RandomCrop} at the same level of variance, showing the effectiveness of semantic-aware localization. Furthermore, with equal semantic information, \textit{ContrastiveCrop} achieves larger variance than \textit{RandomCrop}, which can be owed to center-suppressed sampling.  

We further visualize the similarity of positive pairs in the training process in Fig.~\ref{fig:pos_sim}. As shown in the figure, adding only semantic-aware localization to \textit{RandomCrop} slightly increases similarity, as localization restrains crops in a smaller operable region. Our \textit{ContrastiveCrop} further incorporates center-suppressed sampling, showing smaller positives similarity than the other two. This indicates positive pairs sampled by \textit{ContrastiveCrop} are harder ones, which are helpful in learning more view-invariant features as suggested in FT \cite{zhu2021improving}. However, different from FT that reduces positives similarity in the feature space, we directly sample harder crops from raw data, while taking a careful consideration of semantic information.

%% file: tab/tab_non_IN.tex



\begin{table*}[ht]
\begin{center}

\normalsize
\begin{tabular}{l|c c|c c|c c|c c}
\toprule
Method & \multicolumn{2}{c |}{CIFAR-10} & \multicolumn{2}{c |}{CIFAR-100} & \multicolumn{2}{c |}{Tiny ImageNet} & \multicolumn{2}{c}{STL-10} \\
 & \textit{R-Crop} & \textit{C-Crop} & \textit{R-Crop} & \textit{C-Crop }& \textit{R-Crop} & \textit{C-Crop} & \textit{R-Crop} & \textit{C-Crop} \\
\hline
SimCLR~\cite{chen2020simple} & 89.63 & \bf 90.08 & 60.30 & \bf 61.91 & 45.19 & \bf 46.21 & 88.95 & \bf 89.53 \\
MoCo~\cite{he2020momentum} & 86.73 & \bf 88.78 & 56.10 & \bf 57.65 & 47.09 & \bf 47.98 & 89.17 & \bf 89.81 \\
BYOL~\cite{grill2020bootstrap} & 91.96 & \bf 92.54  & 63.75 & \bf 64.62  & 46.08 & \bf 47.23 & 91.84 & \bf 92.42 \\
SimSiam~\cite{chen2020exploring} & 90.96 & \bf 91.48 & 64.79 & \bf 65.82 & 43.03 & \bf 44.54 & 89.39 & \bf 89.83\\
\bottomrule
\end{tabular}

\caption{Linear classification results for different contrastive methods and datasets. \textit{R-Crop} and \textit{C-Crop} mean \textit{RandomCrop} and \textit{ContrastiveCrop}, respectively. We adopt ResNet-18 as the base model and reproduce all the methods with a unified training setup as described in Sec. \ref{sec:impl}.}
\label{tab:non_IN}
\vspace{-1.5em}
\end{center}
\end{table*}

%% file: sec/exp_v2.tex
\section{Experiments}
In this section, we conduct extensive experiments with popular contrastive methods on a variety of datasets, to demonstrate the effectiveness and generality of our method. We first introduce the datasets and contrastive methods in Sec.~\ref{sec:ds}. Sec.~\ref{sec:impl} describes the implementation details. We then evaluate our method with the common linear evaluation protocol in Sec.~\ref{sec:linear}. Results of ablation experiments are shown in Sec.~\ref{sec:abl}. Finally, Sec.~\ref{sec:downstream} presents transfer performance on downstream object detection and segmentation tasks.

\subsection{Datasets \& Baseline Approaches} \label{sec:ds}
We perform evaluation of our method with state-of-the-art unsupervised contrastive methods, on a wide range of datasets. The datasets include \textbf{CIFAR-10/CIAFR-100} \cite{krizhevsky2009learning}, \textbf{Tiny ImageNet}, \textbf{STL-10} \cite{coates2011analysis} and \textbf{ImageNet}~\cite{deng2009imagenet}. Generally, these datasets are built for object recognition and the images contain iconic view of objects. The baseline contrastive methods include SimCLR~\cite{chen2020simple}, MoCo V1 \& V2~\cite{he2020momentum, chen2020improved}, BYOL~\cite{grill2020bootstrap} and SimSiam~\cite{chen2020exploring}.

\subsection{Implementation Details} \label{sec:impl}
Our \textit{ContrastiveCrop} aims to make better views for contrastive learning, which is agnostic to self-supervised learning frameworks and their related training components, such as backbone networks, losses, optimizers, \etc. Thus, we strictly keep the same training setting when making comparison. Larger gains could be expected with further hyper-parameter tuning, which is not the focus of this work.

For small datasets (\ie, CIFAR-10/100, Tiny ImageNet and STL-10), we use the same training setup in \textit{all} experiments. At the pre-train stage, we train ResNet-18 \cite{res} for 500 epochs with a batch size of 512 and a cosine-annealed learning rate of 0.5. The linear classifier is trained for 100 epochs with initial learning rate of 10.0 multiplied by 0.1 at 60\textit{th} and 80\textit{th} epochs.

For experiments on ImageNet, we adopt ResNet-50 as the base model. Pre-train settings of MoCo V1, MoCo V2 and SimSiam exactly follow their original works. We reproduce SimCLR with a smaller batch size of 512 and cosine-annealed learning rate of 0.05. We adopt the same setting as in \cite{he2020momentum} for training the linear classifier for all baseline methods.

\begin{figure}[htp]
    \centering
    \begin{subfigure}{0.23\textwidth}
        \includegraphics[width=\linewidth, trim=0cm 0cm 1cm 1.2cm, clip]{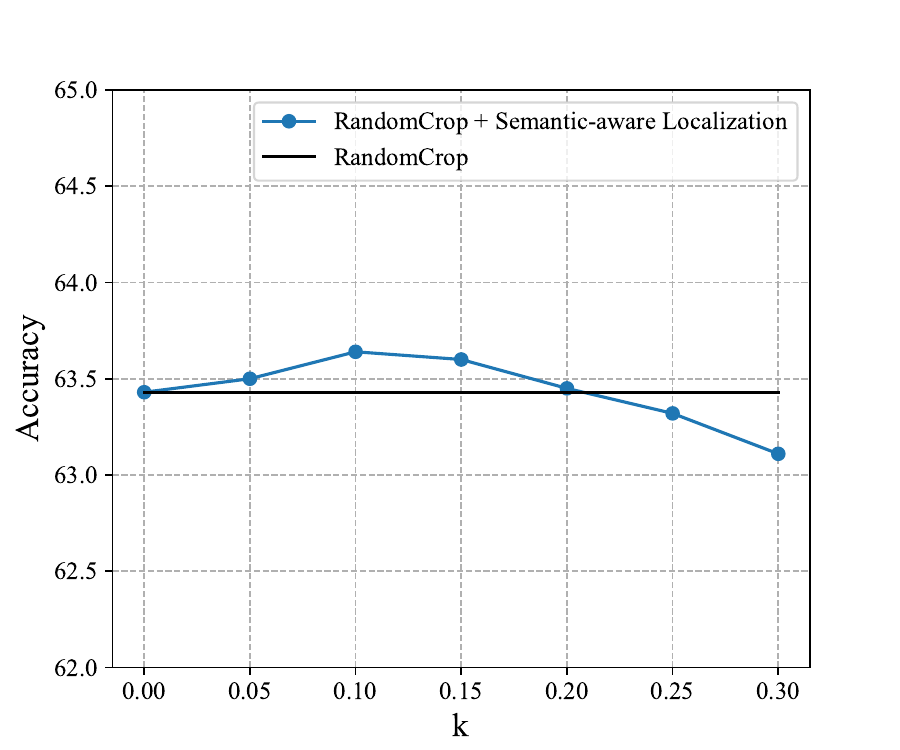}
        \caption{Accuracy \textit{v.s.} $k$}
        \label{fig:acc_k}
    \end{subfigure}
    \begin{subfigure}{0.23\textwidth}
        \includegraphics[width=\linewidth, trim=0cm 0cm 1cm 1.2cm, clip]{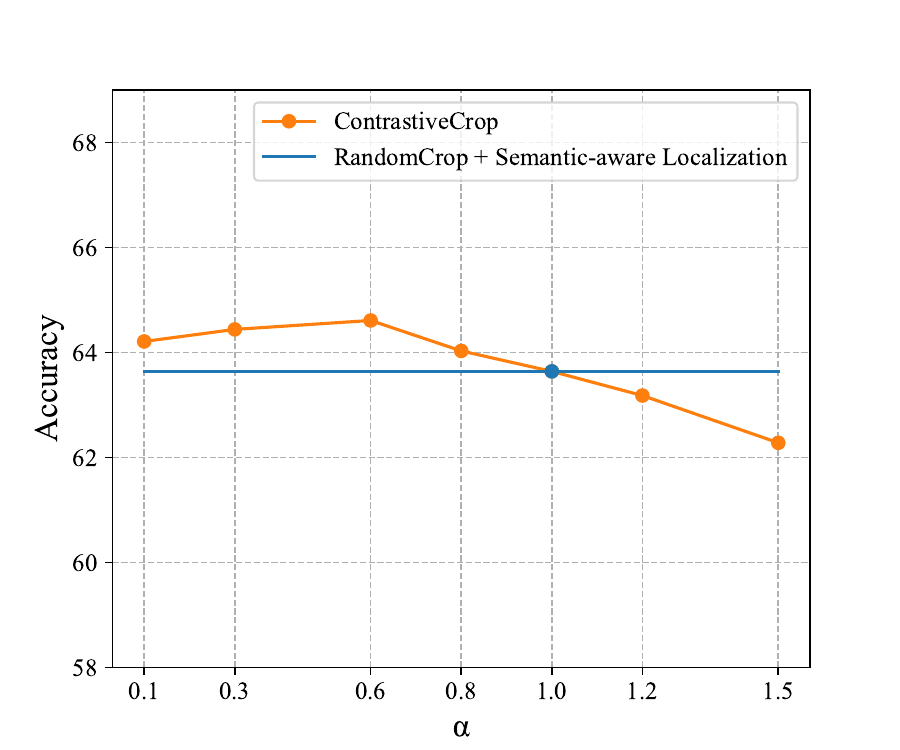}
        \caption{Accuracy \textit{v.s.} $\alpha$}
        \label{fig:acc_alpha}
    \end{subfigure}
\caption{Ablation results on IN-200 w.r.t. $k$ and $\alpha$. Subfigure (a) compares \textit{RandomCrop} baseline (the black plot) with \textit{RandomCrop + Semantic-aware Localization} (the blue plot). In subfigure (b), we fix the best $k=0.1$ for localization (the blue plot) and compare it with \textit{ContrastiveCrop} to study the influence of different $\alpha$.}
\label{fig:k_alpha}
\end{figure}

For our method, we set $k = 0.1$ for the threshold of activations and $\alpha = 0.6$ for sampling. Localization boxes are updated at a frequency of 20\% (\ie, 4 updates in total, except the last epoch), which adds negligible extra training overhead; \textit{RandomCrop} is applied before the first update to collect global information, as described in Sec.~\ref{sec:box}. All the experiments are conducted on an 8-GPU server. We use SGD optimizer with momentum of 0.9, weight decay of $10^{-4}$ and $0$ for pre-train and linear evaluation, respectively.

\subsection{Linear Classification} \label{sec:linear}
In this section, we verify our method with linear classification following the common protocol. We freeze pre-trained weights of the encoder and train a  supervised linear classifier on top of it. Top-1 classification accuracy results on the validation set are reported.

\paragraph{Results on CIFAR-10/100, Tiny ImageNet and STL-10.} 
Our results on these small datasets are shown in Tab.~\ref{tab:non_IN}. With the same training setup for \textit{all} experiments, \textit{ContrastiveCrop} consistently improves baseline methods by at least $0.4\%$. Results show that the proposed method is generic and does not require heavy parameter tuning. The localization boxes are updated at a frequency of 20\% in the training process (\ie, 4 times in total, except the last epoch), adding negligible training overhead.

\paragraph{Results on ImageNet.} The results of ImageNet are two-part: 1) standard ImageNet-1K (IN-1K), which is used for pre-training. 2) IN-200, which consists of 200 random classes of IN-1K and is used for ablation experiments. As shown in Tab.~\ref{tab:IN}, our method outperforms \textit{RandomCrop} with SimCLR, \mbox{MoCo V1}, \mbox{MoCo V2}, SimSiam on IN-1K by $0.25\%$, $1.09\%$, $0.49\%$ and $0.33\%$, respectively. A larger improvement is seen on IN-200. The consistent gain over baseline methods shows the effectiveness and generality of \textit{ContrastiveCrop} for contrastive methods.

\input{tab/tab_IN}

\subsection{Ablation Studies} \label{sec:abl}
In ablation studies, we investigate the semantic-aware localization module and center-suppressed sampling respectively. We also study the effect of \textit{ContrastiveCrop} when it is combined with different transformations. We conduct experiments with MoCo V2 and ResNet-50, and report the linear classification results on IN-200. 

\paragraph{Semantic-aware Localization.} In our method, the unsupervised semantic-aware localization serves as a guidance to make crops. We study the influence of $k$ that determines the scale of the localization box, with a larger $k$ leading to a smaller box. We also make comparison with \textit{RandomCrop} that does not use localization (\ie, $k=0$). Experimental results are shown in Fig.~\ref{fig:acc_k}. One can find that using localization box outperforms \textit{RandomCrop} baseline (the black plot) within a range from 0.05 to 0.2. This shows the effectiveness of largely removing false positives. However, as $k$ increases over $0.25$, the performance starts to fall quickly. We suggest the reason is smaller bounding boxes dramatically reduce variance of views, making it trivial to learn discriminative features.


\input{tab/tab_freq}

We also study the effect of update frequency of localization boxes in Tab.~\ref{tab:freq}. It shows that only one update in the middle of training (\ie, $50\%$) could outperform \textit{RandomCrop} baseline (\ie, $0\%$). A larger improvement is seen in a range of $10\% \sim 30\%$ where there are more updates. These results show that our method could work well for different update frequencies.

\input{tab/tab_dst}

\paragraph{Center-suppressed Sampling.} In this work, we use $\beta$ distribution for the center-suppressed sampling, which allows to control its variance with different $\alpha$. Here, we investigate the impact of different variance by iterating over multiple $\alpha$. Results are shown in Fig. \ref{fig:acc_alpha} with $k = 0.1$ for localization. When $\alpha < 1$, our \textit{ContrastiveCrop} consistently outperforms \textit{RandomCrop} with localization, showing the effect of center-suppressed sampling. We also study $\alpha > 1$ that has a smaller variance than uniform distribution (\ie, $\alpha=1$). A drop in accuracy is observed with $\alpha > 1$. This indicates that larger variance of crops is required for better contrast.

\paragraph{\textit{ContrasitveCrop} with Other Transformations.} To further compare the effect of \textit{ContrastiveCrop} and \textit{RandomCrop}, we study their combinations with other image transformations. Here, we choose the transformations used in MoCo V2 \cite{chen2020improved}, including \textit{Flip}, \textit{ColorJitter}, \textit{Grayscale} and \textit{Blur}. The ablation results are shown in Tab.~\ref{tab:trans}. In case all other transformations are removed, \textit{ContrastiveCrop} is $0.4\%$ higher than \textit{RandomCrop}, which is a direct evidence of its superiority. Moreover, with only one extra transformation, \textit{ContrastiveCrop} outperforms \textit{RandomCrop} by $0.3\% \sim 0.8\%$. The largest gap of $1.2\%$ is achieved when all of the transformations are incorporated, which indicates that the potential of \textit{ContrastiveCrop} can be larger exploited with further color transformations. Additionally, these results show that our \textit{ContrastiveCrop} is compatible and orthogonal to other transformations.

\input{tab/tab_trans}

\subsection{Downstream Tasks} \label{sec:downstream}
In this section, we measure the transferability of our method on the object detection and instance segmentation task. 
Following previous works \cite{he2020momentum, zhu2021improving}, we pre-train \mbox{ResNet-50} on \mbox{IN-1K} for 200 epochs. For downstream tasks, we use PASCAL VOC \cite{everingham2010pascal_VOC} and COCO \cite{lin2014microsoft_coco} as our benchmarks and we adopt the same setups as in MoCo's detectron2 codebase \cite{wu2019detectron2}. All layers of pre-trained models are fine-tuned end-to-end at target datasets.

\paragraph{PASCAL VOC Object Detection.} Following \cite{he2020momentum}, we use Faster R-CNN \cite{faster} with a backbone of R50-C4 \cite{he2017mask} as the detector. We fine-tune the model on the \texttt{trainval2007+2012} split and evaluate on the VOC \texttt{test2007}. The results are present in Tab.~\ref{tab:dst}. Compared with MoCo V1 baseline, our method achieves a consistent improvement of +0.2AP,  +0.2AP$_{50}$ and +0.4AP$_{75}$.

\paragraph{COCO Object Detection/Instance Segmentation.} The model for both detection and segmentation is Mask R-CNN \cite{he2017mask} with R50-C4 backbone. We fine-tune 90K iterations on the \texttt{train2017} set and evaluate on \texttt{val2017}. As shown in Tab. \ref{tab:dst}, the proposed \textit{ContrastiveCrop} achieves superior performance in all metrics.

%% file: tab/tab_IN.tex

\begin{table}[ht]
    \centering
    \small
    \begin{tabular}{l|c|c|c|c}
        \toprule
        \multirow{2}{*}{\bf Method} & 
        \multirow{2}{*}{\bf Arch.} &
        \multirow{2}{*}{\bf Epoch} & \bf IN-200 & \bf IN-1K \\
        & & & \bf Top-1 & \bf Top-1 \\
        \midrule
        SimCLR & R50 & 100 & 62.14 & 61.60 \\
        \rowcolor{graylite} SimCLR + Ours & R50 & 100 & \bf 63.08 & \bf 61.85 \\
        \hline
        MoCo V1 & R50 & 100& 64.52 & 57.25 \\
        \rowcolor{graylite} MoCo V1 + Ours & R50 & 100 & \bf 65.80 & \bf 58.34 \\
        \hline
        MoCo V2 & R50 & 100 & 63.43 & 64.40 \\
        \rowcolor{graylite} MoCo V2 + Ours & R50 & 100 & \bf 64.61 & \bf 64.89 \\
        \hline
        SimSiam & R50 & 100 & 62.89 & 65.62 \\
        \rowcolor{graylite} SimSiam + Ours & R50 & 100 & \bf 63.54 & \bf 65.95 \\
        \bottomrule
    \end{tabular}
    \caption{Comparison of \textit{RandomCrop} and our \textit{ContrastiveCrop} with linear classification results on IN-200 and IN-1K. Models are pre-trained for 100 epochs, with the same training setup within a method for fair comparison.}
    \label{tab:IN}
\end{table}

%% file: tab/tab_freq.tex
\begin{table}[h]
    \centering
    \begin{tabular}{c|c|c|c|c|c}
         Freq. & 0\% & 10\% & 20\% & 30\% & 50\% \\
         \hline
         Acc. (\%) & 63.43  & 64.40 &  64.61  &  64.40  &  64.11
    \end{tabular}
    \caption{Linear classification accuracy w.r.t different update frequencies of localization boxes. $0\%$ means no update (\textit{i.e.}, \textit{RandomCrop} baseline) and $50\%$ means one update in the right middle of training. \textit{RandomCrop} is applied before the first update.}
    \label{tab:freq}
\end{table}

%% file: tab/tab_dst.tex
\begin{table*}
\centering
	\resizebox{0.999\textwidth}{!}{
	\begin{tabular}{l|c|c c c |ccc| ccc }
		\toprule
		\multirow{2}{*}{Pre-train}& IN-1K &\multicolumn{3}{c|}{VOC detection}  & \multicolumn{3}{c|}{COCO instance seg.} & \multicolumn{3}{c}{COCO detection}  \\ 
		& Top-1 & AP   & AP$_{50}$   & AP$_{75}$ & AP$^{mk}$ & AP$^{mk}_{50}$ & AP$^{mk}_{75}$ & AP$^{bb}$ & AP$^{bb}_{50}$ & AP$^{bb}_{75}$  \\
		\midrule
Random init&- 		& 33.8  & 60.2  & 33.1   & 29.3  & 46.9 & 30.8 & 26.4 & 44.0  & 27.8    \\
		Supervised & 76.1	& 53.5 	& 81.3  & 58.8   & 33.3  & 54.7 & 35.2 & 38.2 & 58.2  & 41.2    \\
		InfoMin \cite{tian2020makes}    & 70.1  	& 57.6	& 82.7  & 64.6   & 34.1  & 55.2 & 36.3 & 39.0 & 58.5  & 42.0    \\
		\midrule
		MoCoV1 \cite{he2020momentum}    & 60.6 	& 55.9  & 81.5  & 62.6   & 33.6  & 54.8 & 35.6 & 38.5 & 58.3  & 41.6    \\
		MoCoV1 + \textit{ContrastiveCrop}   & \bf 63.0 	& \bf 56.1  & \bf 81.7	& \bf 63.0   & \bf 33.9  & \bf 55.2  & \bf 36.1 & \bf 38.8 & \bf 58.5  & \bf 41.9     \\
		\midrule
        MoCoV2 \cite{chen2020improved}    & 67.5	& 57.0  & 82.4  & 63.6  & 34.2 & 55.4  & 36.2 & 39.0  & 58.6 & 41.9 \\
        MoCoV2 + \textit{ContrastiveCrop}   & \bf 67.8 	& \bf 57.3  & \bf 82.5	& \bf 63.8  & \bf 34.5  & \bf 55.5  & \bf 36.4 & \bf 39.2 & \bf 58.8  & \bf 42.2     \\
		\bottomrule
	\end{tabular}}
    \caption{Fine-tuning results on PASCAL VOC detection and COCO detection and instance segmentation. All models are pre-trained for 200 epochs on ImageNet-1K. On VOC, the training and evaluation sets are \texttt{trainval2007+2012} and \texttt{test2007}, on COCO are the \texttt{train2017} and \texttt{val2017}. All models are fine-tuned for 24K iterations on VOC and 90K on COCO.}
\label{tab:dst}
\end{table*}

%% file: tab/tab_trans.tex
\begin{table}[ht]
    \centering
    \small
    \begin{tabular}{ccc|ccc|}
	    \toprule
	    \textit{Flip} & \textit{ColorJitter} + \textit{Grayscale} & \textit{Blur} & \textit{R-Crop} & \textit{C-Crop} \\
	    \midrule
	    \checkmark & \checkmark & \checkmark & 63.4 & \bf 64.6 \\
	    \checkmark &            &            & 50.4 & \bf 50.9 \\
	               & \checkmark &            & 60.6 & \bf 61.4 \\
	               &            & \checkmark & 44.9 & \bf 45.2 \\
	               &            &            & 45.5 & \bf 45.9 \\
        \bottomrule
    \end{tabular}
    \caption{Ablation of other transformations used in MoCo V2. We combine \textit{ColorJitter} and \textit{Grayscale} as a single color transformation. \textit{R-Crop} and \textit{C-Crop} denote \textit{RandomCrop} and \textit{ContrastiveCrop},  respectively. The results are from ResNet-50 pre-trained on IN-200 for 100 epochs.}
    \label{tab:trans}
\end{table}

%% file: sec/conclusion.tex
\section{Discussion and Conclusion}
In this work, we propose \textit{ContrastiveCrop}, that is tailored to make better contrastive views for Siamese representation learning. \textit{ContrastiveCrop} adopts semantic-aware localization to avoid most false positives and applies the center-suppressed sampling to reduce trivial positive pairs. We innovatively take semantic information into account when transforming a sample, and thoroughly investigate the suitable variance for contrastive learning. We have shown the effectiveness and generality of our method through extensive experiments with state-of-the-art contrastive methods including SimCLR, MoCo, BYOL and SimSiam. Finally, we hope this work could inspire future research of positives designing, considering its significant role in contrastive learning.





\section{Acknowledge}
This research is supported by the National Research Foundation, Singapore under its AI Singapore Programme (AISG Award No: AISG2-PhD-2021-08-008). We thank Google TFRC for supporting us to get access to the Cloud TPUs. We thank CSCS (Swiss National Supercomputing Centre) for supporting us to get access to the Piz Daint supercomputer. We thank TACC (Texas Advanced Computing Center) for supporting us to get access to the Longhorn supercomputer and the Frontera supercomputer. We thank LuxProvide (Luxembourg national supercomputer HPC organization) for supporting us to get access to the MeluXina supercomputer.